%% file: grammar-design.tex
\documentclass[sigconf]{acmart}

\usepackage{booktabs} % For formal tables
\usepackage{verbatim}
\usepackage{graphicx}
\usepackage{subcaption}
\setcopyright{rightsretained}
\acmDOI{10.1145/nnnnnnn.nnnnnnn} % To be updated after completing copyright process
\acmISBN{978-x-xxxx-xxxx-x/YY/MM} % To be updated after completing copyright process
\acmConference[GECCO '22]{The Genetic and Evolutionary Computation Conference 2022}{July 9--13, 2022}{Boston, USA}
\acmYear{2022}
\copyrightyear{2022}

\acmPrice{15.00}

\DeclareUnicodeCharacter{2217}{*}

\begin{document}
\title{Initialisation and Grammar Design in Grammar-Guided
  Evolutionary Computation}
\titlenote{∗An abridged version of this paper appears in~\cite{dick2022grammars}}
\author{Grant Dick}
\affiliation{%
  \institution{Department of Information Science\\University of Otago}
  \city{Dunedin} 
  \country{New Zealand} 
}
\email{grant.dick@otago.ac.nz}
\author{Peter A. Whigham}
\affiliation{%
  \institution{Department of Information Science\\University of Otago}
  \city{Dunedin} 
  \country{New Zealand} 
}
\email{peter.whigham@otago.ac.nz}
% \author{Anon}
% \affiliation{%
%   \institution{Inst 1\\Inst 2}
%   \city{City} 
%   \country{Country}
% }
% \email{abc@xyz}

\renewcommand{\shortauthors}{G. Dick and P. A. Whigham}

\begin{abstract}
  \input{abstract}
\end{abstract}




\maketitle

\section{Introduction}
Grammars provide a convenient and powerful mechanism to define the
space of possible solutions for a range of problems, and the
incorporation of grammars into evolutionary computation has a history
spanning over three
decades~\cite{cramer1985representation,whigham1995grammatically,o2001grammatical}. While
a number of methods exist for grammar-guided evolutionary computation
(GGEC), most research into the use of grammars in evolution has
focused upon grammatical evolution (GE), which recently marked twenty
years of research and development\cite{ryan2018introductionC}. While
GE has been applied to a range of problems from regression to
evolutionary design and hyperheuristics, a significant portion of GE
research has focused on understanding the interaction between its
linear representation, genotype-to-phenotype mapping, and the
polymorphic treatment of codons within its mapping
process. Traditionally, this mapping from linear representation to
non-linear phenotype has drawn inspiration from molecular biology, and
has been argued to offer benefits of degeneracy, neutrality and
diversity preservation. However, previous work has questioned the
utility of GE's linear representation; specifically, it is argued that
the linear representation leads to low locality in search, and poor
performance characteristics that resemble random search on some
problems~\cite{eurogp06:RothlaufOetzel,whigham2015examining,whigham2017mapping}. It is argued that other mechanisms of GGEC,
such as context-free grammar genetic programming (CFG-GP) do not
demonstrate such limitations, and are thus stronger choices for
evolutionary computation that is influenced by grammars.

Understandably, the response to these criticisms of GE has been an
increase in work to understand the underlying characteristics of GE
search, and to design extensions to GE that improve its search
characteristics. Some of this work has explored modifications of the
linear representation so that it preserves structural information
about the grammar during
evolution~\cite{conf/cec/HarperB05,lourencco2016unveiling,bartoli2018weighted}. Other
work has focused on the initialisation process used in GE, and the
design of the grammars that are used to describe the problem, and the
resulting extensions to GE demonstrate improved performance on a range
of benchmark
problems~\cite{harper2010ge,nicolau2017understanding,nicolau2018understanding}. In
this latter body of work, the improvements to GE are typically
examined without comparison to other grammar-based methods, so the
relative merits and utility of these improvements remaining largely
unknown. The goal of this paper is to take some of the results of
previous work and compare them to the CFG-GP and random search that
has been subjected to the same alterations of initialisation and
grammar design. After testing on a range of benchmark problems,
results suggest that CFG-GP is largely insensitive to the
initialisation method used and is able to adequately recover from a
poor initial population. The results align with previous work in that,
for several of the examined problems, GE's performance is roughly
equivalent to random search using the same initialisation routines and
grammar designs. Additionally, on the two examples where CFG-GP
performs relatively poorly, it is shown that potential limitations in
the behaviour of CFG-GP may be overcome through more effective
parameter tuning, rather than relying upon elaborate modification of
the grammar to match the representation. The findings of this paper
suggest that future work should place greater emphasis on both the
understanding and application of CFG-GP, and that the design of
grammars should focus on more effective expression of the problem
itself in order to make good solutions easier to discover.

The remainder of this paper is structured as follows:
\S\ref{sec:background} briefly examines recent developments in GGEC
and GE; \S\ref{sec:setup} describes the experimental framework used in
this paper; \S\ref{sec:results} presents the results of
experimentation over several benchmark functions using a range of
initialisation methods and grammars; \S\ref{sec:discussion} assesses
the results of the previous section and demonstrates a simple
extension to CFG-GP that appears to ease its application to two of the
benchmarks; finally, \S\ref{sec:conc} concludes the paper and suggests
areas of future research.

\section{Recent Work in Grammar-Guided Evolutionary Computation}\label{sec:background}
The use of grammars in evolutionary computation has a long-established
track record, and previous work provides a thorough review of this
research~\cite{mckay2010grammar}. More recent work on GGEC has examined potential
limitations of grammatical evolution, focusing on locality and
problems with the genotype-to-phenotype mapping. Indeed, some work has
gone so far as to suggest that GE's behaviour on many problems is
roughly equivalent to random search. In response to this criticism,
new flavours of grammatical evolution have been developed, aimed at
creating more direct and structure preserving
representations~\cite{lourencco2016unveiling,bartoli2018weighted}. The resulting solutions incorporate
derivation tree information into the representation, creating
derivatives of GE that more closely resemble CFG-GP. Other work has
provided greater understanding of the mapping process of linear
genotypes and has lead to developments in grammar design and
initialisation to improve the performance of standard GE~\cite{nicolau2017understanding,nicolau2018understanding}.

\subsection{Initialisation}
The limitations of straight random initialisation of genotypes was
identified early in the history of grammatical evolution. Subsequent
work in GE focused on developing initialisation methods that operated
in the phenotype space and back-fitted the resulting derivations into
a linear representation~\cite{ryan+azad:2003:gecco:workshop}. The resulting `sensible'
initialisation essentially re-implemented the ramped-half-and-half
method from standard genetic programming into GE, and presented the
first work that begins to blur the separation of genotype and
phenotype in GE.\footnote{Although it is essentially a grammar-aware version of ramped-half-and-half initialisation, we use the name `sensible initialisation' throughout this paper to remain consistent with previous work~\cite{ryan+azad:2003:gecco:workshop,nicolau2017understanding}.}

Following this work on sensible initialisation, work in GE has
examined the use of probabilistic tree creation (PTC) methods for
initialisation. Like sensible initialisation, PTC works in the
phenotype (i.e., derivation tree) space and in the context of GE the
resulting solution is then back-mapped into a linear
representation. In PTC, the leaves of a tree are developed in a more
breadth-like manner than the depth-first approach typically used in
genetic programming~\cite{luke2000two}. Results in grammatical
evolution have suggested that a strongly-typed version of PTC, named
PTC2, can significantly improve the performance of
GE~\cite{harper2010ge,nicolau2017understanding}. This is particularly
interesting, as previous work has suggested that neither PTC nor PTC2
greatly improve upon standard initialisation routines in genetic
programming~\cite{luke2001survey}. Theories to explain the success of
PTC2 in grammatical evolution remain elusive, while its behaviour in
other search mechanisms (such as random search or CFG-GP) remains
unknown.

\subsection{Grammar Design}
The purpose of a grammar is to define the structure and syntactical
properties of the language used to express solutions to a given
problem. Ideally, grammar design should focus on efficient and elegant
expression of language, and preserving modularity in solutions. From
the context of evolutionary search, the design of the grammar should
be agnostic to the representation used to search the space of
solutions. However, it was identified early in GE research that the
design of the grammar itself plays a significant role in the
effectiveness of GE on a given problem~\cite{nicolau2004automatic}. The polymorphic
nature of the genotype-to-phenotype mapping in GE means that careful
design of grammar productions is required to avoid search
bias.\footnote{Here, {\em bias} refers to tendency towards certain
  productions in the grammar due to the nature of the linear
  representation, rather than the more usual definition of {\em
    grammatical bias} domain knowledge is incorporated into search
  through special productions.} Indeed, even small changes in grammar
design can have significant impact on search, making comparisons of
results on the same problem across different papers
difficult~\cite{robilliard2005santa}.

\begin{table*}[t]
  \centering
  \caption{The major parameter settings used throughout all experiments.}\label{tab:parameters}
  \begin{tabular}{lrrr}
    \toprule
              & \multicolumn{3}{c}{Setting} \\
    Parameter & GE & Random Search & CFG-GP \\    
    \midrule
    Population Size & 500 & 25050 & 500\\
    Generations & 50 & 1 & 50 \\
    Elitism & 1\% & - & 1\% \\
    Tournament Size & 1\% & - & 1\% \\
    Crossover Rate & 0.5 & - & 0.9 \\
    Mutation Rate & 1 per individual & - & 0.1 \\
    Random Initialisation & 31 codons & 31 codons & max depth 6 \\
    Sensible Initialisation & max depth 6 & max depth 6 & max depth 6 \\
    PTC2 Max Expansions & 31 & 31 & 31 \\
    Maximum Tree Depth & - & - & 17 \\
    Maximum Mutation Depth & - & - & 4 \\
    Maximum Wrapping Events & 0 & 0 & - \\
    \bottomrule
  \end{tabular}
\end{table*}

To reduce the mismatch between linear representations and grammar
productions, recent work has suggested the following steps~\cite{nicolau2018understanding}:
\begin{enumerate}
\item {\em Balancing} --- grammars should be designed to ensure equal
  probability of termination or expansion during
  mapping. Where an imbalance between termination and
  expansion exists, the rarer productions can be duplicated to provide
  this balance.
\item {\em Unlinking} --- the use of modulo arithmetic in GE's mapping
  process creates linkages between the productions of different
  non-terminals, meaning that certain codon values in the
  representation are only able to represent specific subsets of
  productions. Unlinking can be done by duplication of
  productions in non-terminals to effectively create a Cartesian
  product of productions indices, resulting in codon values that are
  potentially more expressive during mapping.
\item {\em Eliminating non-terminals} --- the non-terminals define the
  structural elements of the grammar, and can be used to create
  modularity in the search, or to form salient end-user documentation
  of the nature of the grammar. However, previous work in GE has
  identified that having excessive non-terminals in the grammar
  interacts with GE's polymorphic interpretation of codons during
  mapping, and may lead to fragile search. The solution
  is to fold the productions of the grammar into a single global
  non-terminal, with the resulting grammar closely resembling that of
  the implicit `closure' grammar used in standard genetic
  programming. This may produce an imbalance of recursive productions,
  but this can be balanced by further duplication of productions that
  lead to termination.
\item {\em Removing grammar biases} --- the nature of the way that
  grammars are designed means that not all terminals are sampled at
  equal rates. Again, this can be remedied through careful alignment
  of production choices and duplication of productions as necessary.
\item {\em Prefix notation} --- certain grammar designs, particularly
  those used in symbolic regression, have the choice of using infix or
  prefix notation. Previous work has noted that using an infix
  notation may have lower locality compared to a prefix notation due
  to the required operator precedence rules.
\item {\em Compromise grammars} --- all of the previously mentioned
  steps may induce changes that cause the size and complexity of the
  grammar to grow exponentially. A useful compromise to
  these steps is to reintroduce a single non-terminal to express
  constants and variables in the problem. This has some negative
  impact on crossover locality, but results in a grammar that is
  significantly smaller and easier to interpret.
\end{enumerate}
In previous work, the grammars that result from these modifications
are labelled g1-g6, with g0 denoting the unmodified grammar that is
typically encountered in previous
work~\cite{nicolau2018understanding}. We adopt this notation in this
paper.

It is interesting to note that the first three proposed remedies are a
consequence of two factors: GE's polymorphic interpretation of codons
during mapping, and GE's lack of structure preserving operators during
search that induces this polymorphic interpretation. Additionally,
these modifications (particularly the steps that eliminate
non-terminals and remove bias) substantially increase the number of
productions in the grammar by multiple orders of magnitude (e.g., from
22 to 3300 in the {\em Keijzer-6} problem that is examined later in
this paper), which may have significant implications on methods that
use grammar productions directly to create and modify solutions (e.g.,
PTC2 initialisation). Finally, it should be noted that the third step
(removal of non-terminals) can only be applied in domains where the
solution lacks modularity, so is precluded from domains where
structure is important (e.g., in the evolution of hyperheuristics~\cite{journals/tec/BurkeHK12}).

\section{Setup}\label{sec:setup}
Previous work has examined GE's behaviour on a number of benchmark
problems using a range of grammar designs and initialisation
methods~\cite{nicolau2018understanding}. To provide a more complete the picture for grammar-guided
evolutionary computation in general, we repeat a number of these
experiments, but this time include random search (where a initialisation
method is run for a given number of times and the best solution find
in all of these samples is returned) and CFG-GP. We reuse problems and algorithm settings from previous work to provide a fair comparison~\cite{nicolau2018understanding,whigham2015examining}. Algorithm parameters used in this study are described in Table~\ref{tab:parameters}: for Sensible
Initialisation and PTC2, the minimum depth and expansion parameter
values were determined by inspecting the productions in the
grammar. We focus on the problems that received the most attention in
the previous study, and whose grammars were fully provided: {\em
  Keijzer-6}, {\em Vladislavleva-4}, and the hard {\em Shape}
problem. Details of these problems can be found in the previous work
and are omitted for space reasons~\cite{nicolau2018understanding}. In
addition, we examine four regression problems used in previous work:
{\em Boston Housing}, {\em Dow Chemical}, {\em Forest Fires}, and {\em
  Tower}. We explored two grammar designs for the regression problems,
a structured grammar (g0) that would be typical of that used in
grammar-guided search, and a `compromise' design (g6) that is informed
by the processes suggested in the previous section. The two grammars
are presented in Figure~\ref{fig:reg-grammars}. Finally, we also
examine the {\em Santa Fe Trail} problem, as it has been the subject
of several examples of previous work on grammar
design~\cite{o1999evolving,robilliard2005santa,harper2010ge,nicolau2012termination}. We
examined four of these grammars from previous work (g0-g3), and these
are presented in Figure~\ref{fig:ant-grammars}. In addition, we
designed a novel grammar (g4) for this problem that was informed by
inspection of the other grammars and the resulting solutions that were
evolved through their use. This final grammar was used to demonstrate
the benefits of grammar design where the emphasis is on clearer and
more effective definition of the language, rather than modification of
the grammar to suit algorithmic requirements. We acknowledge that use
of the {\em Santa Fe Trail} problem specifically as a benchmark is
dubious, but we include it here primarily as a tool to examine
potential issues (and subsequent remedies) with the use of CFG-GP.

\begin{figure}[t]
  \begin{subfigure}{\linewidth}
\begin{verbatim}
<prog>  ::= <expr>
<expr>  ::= <op>  <expr>  <expr> | <trans> <expr>
          | <erc> | <var>
<op>    ::= + | * | - | pdiv
<trans> ::= sin | cos | sqr | log
<erc>   ::= <GECodonValue{-1.000 : 1.000 : 0.001}>
<var>   ::= x1 | x2 | ... | xn
\end{verbatim}
    %%\vspace{-1.5em}
    \caption{g0: structured grammar}
  \end{subfigure}\\[1em]
  \begin{subfigure}{\linewidth}
\begin{verbatim}
<expr>  ::= + <expr> <expr> | <term>
          | * <expr> <expr> | <term>
          | - <expr> <expr> | <term>
          | pdiv <expr> <expr> | <term>
          | sin <expr> | cos <expr>
          | sqr <expr> | log <expr>
<term>  ::= <GECodonValue{-1.000 : 1.000 : 0.001}>
          | x1 | x2 | ... | xn
\end{verbatim}
    %%\vspace{-1.5em}
    \caption{g6: `compromise' grammar with reduced non-terminal set}
  \end{subfigure}
  \caption{The grammars used for the four regression problems. The
    {\tt <GECodonValue\{-1.000 : 1.000 : 0.001\}>} terminal has a
  different interpretation between GE-based and derivation tree-based
  methods. For GE, this
  takes the codon value and maps it into a numeric constant in the
  interval $[-1,1]$, as in previous work~\cite{nicolau2006introducing}. In tree-based
  methods, {\tt <GECodonValue\{-1.000 : 1.000 : 0.001\}>} generates a
  random constant in the interval $[-1,1]$ during tree construction,
  and embeds this value as a terminal in the derivation tree. In both
  cases, the resulting constant is rounded to three decimal places.}
  \label{fig:reg-grammars}
\end{figure}

The source code used in all experiments is
available online.\footnote{URL available for final publication.}

\section{Results}\label{sec:results}
\begin{figure}[t]
  \centering
  \includegraphics[width=\linewidth]{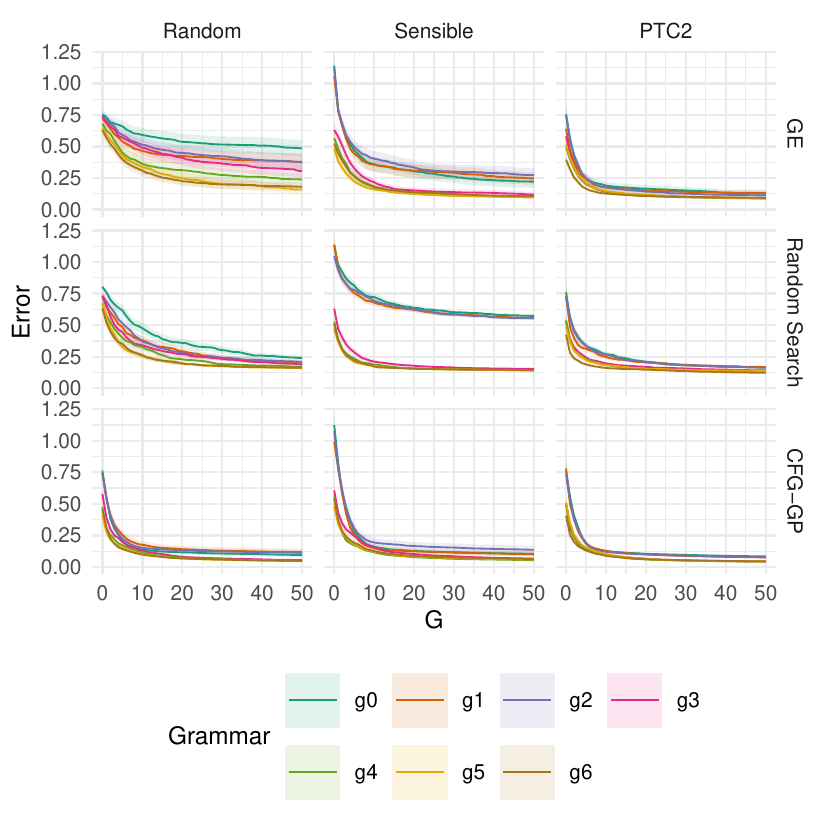}
  \caption{Fitness evolution over generations for the {\em Keijzer-6} problem.} \label{keijzer6}
\end{figure}

\begin{figure}
  \centering
  \includegraphics[width=\linewidth]{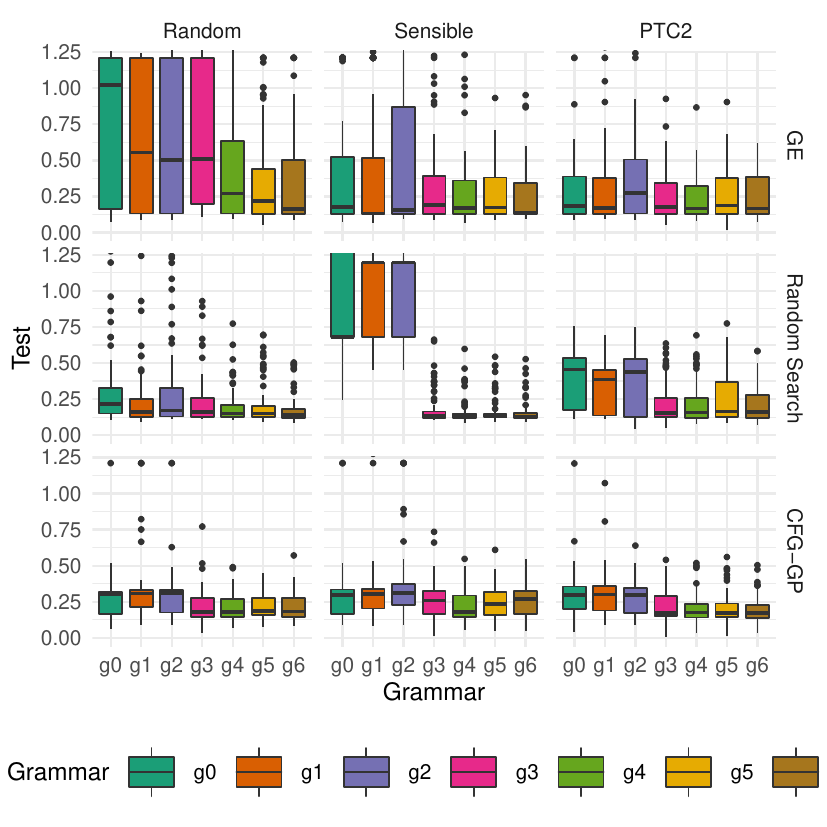}
  \caption{Test performance of best solution in final generation for the {\em Keijzer-6} problem.} \label{keijzer6-test}
\end{figure}

\begin{figure}[t]
  \centering
  \includegraphics[width=\linewidth]{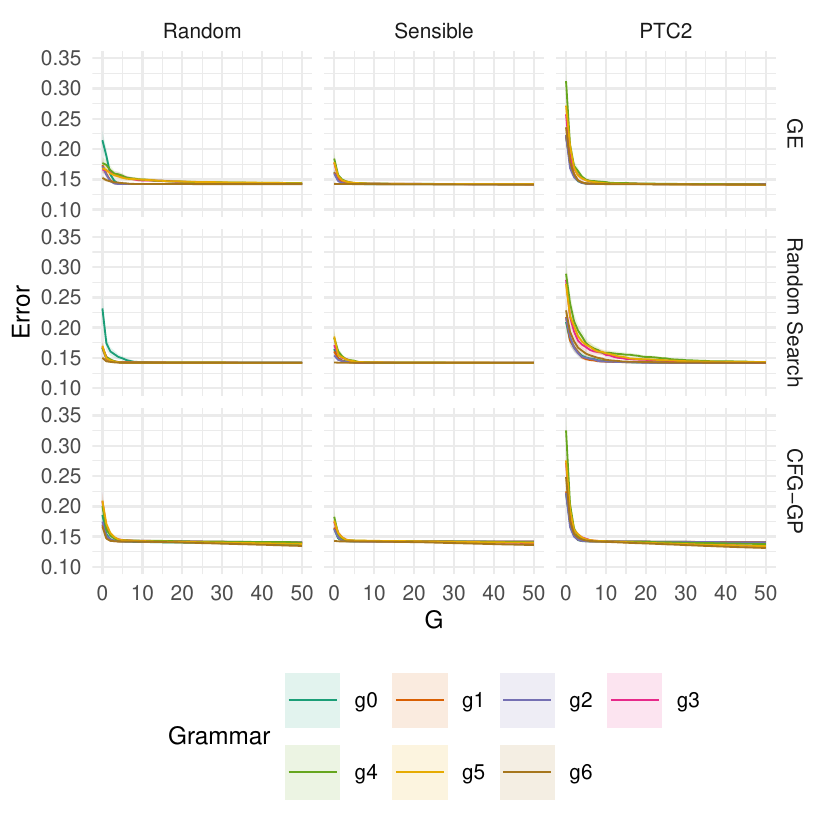}
  \caption{Fitness evolution over generations for the {\em Vladislavleva-4} problem.} \label{vladislavleva4}
\end{figure}

\begin{figure}[t]
  \centering
  \includegraphics[width=\linewidth]{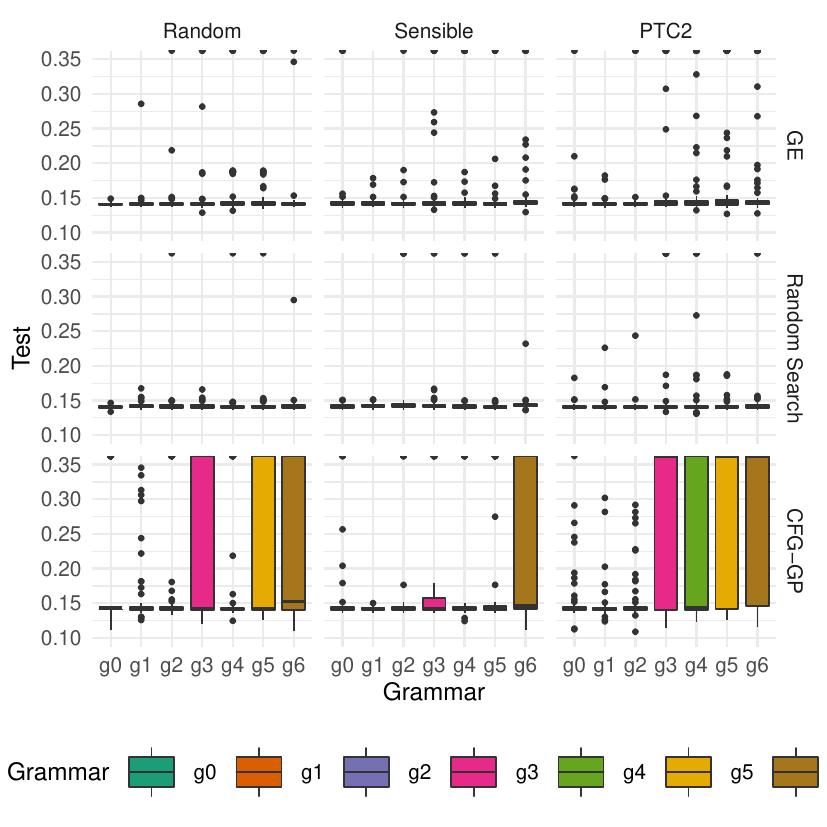}
  \caption{Test performance of best solution in final generation for the {\em Vladislavleva-4} problem.} \label{vladislavleva4-test}
\end{figure}

\begin{figure}[t]
  \centering
  \includegraphics[width=\linewidth]{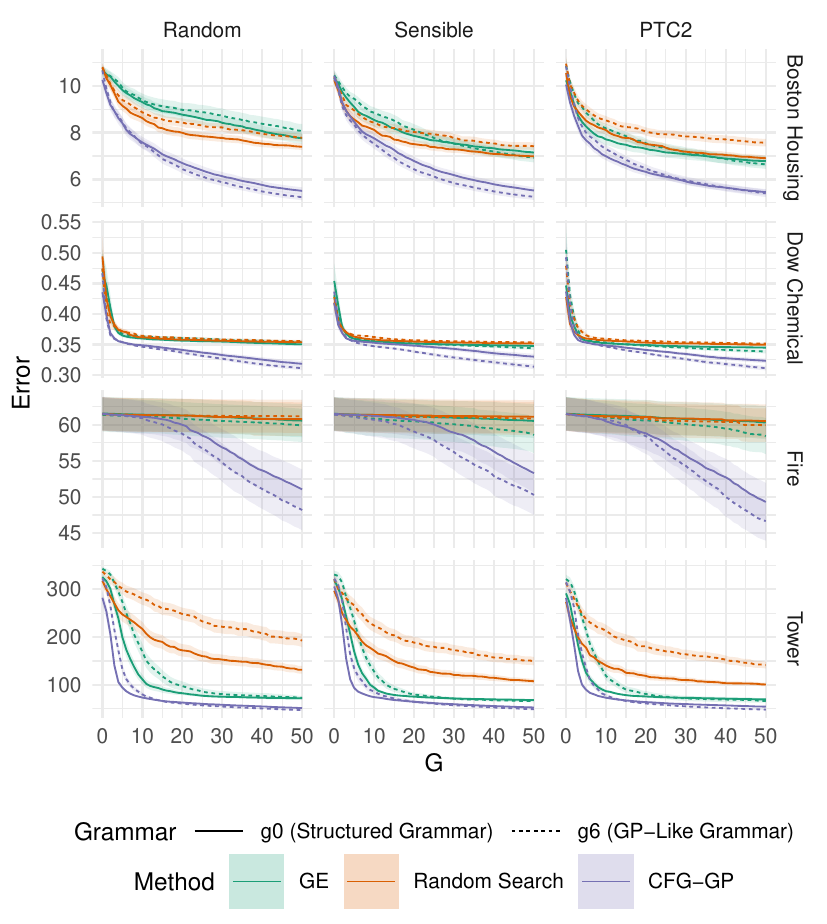}
  \caption{Fitness evolution over generations for the four additional symbolic regression problems.} \label{regression}
\end{figure}

\begin{figure}[t]
  \centering
  \includegraphics[width=\linewidth]{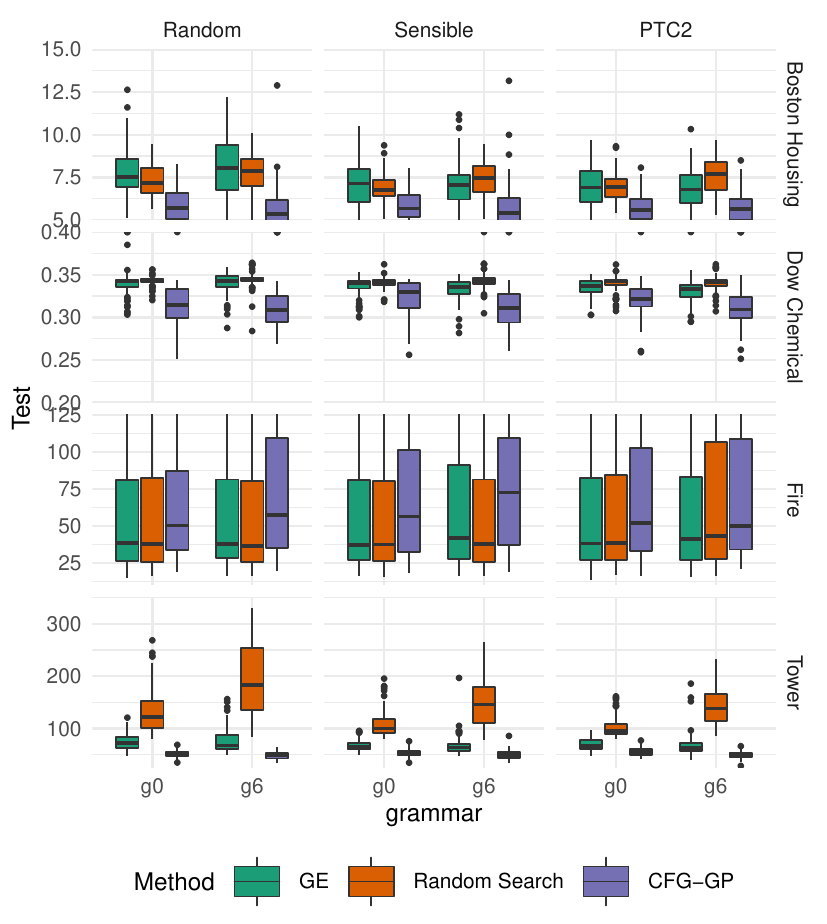}
  \caption{Out of sample test errors for the four additional symbolic regression problems.} \label{regression-test}
\end{figure}

\begin{figure}[t]
  \centering
  \includegraphics[width=\linewidth]{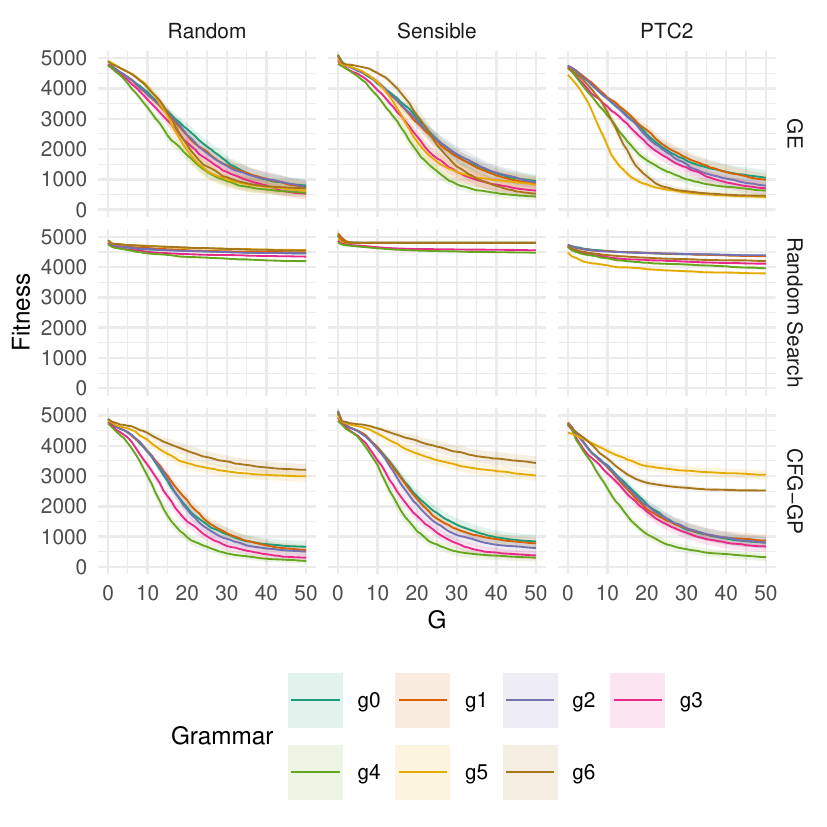}
  \caption{Fitness evolution over generations for the {\em Shape} problem.} \label{shape}
\end{figure}

\begin{figure}[t]
  \centering
  \includegraphics[width=\linewidth]{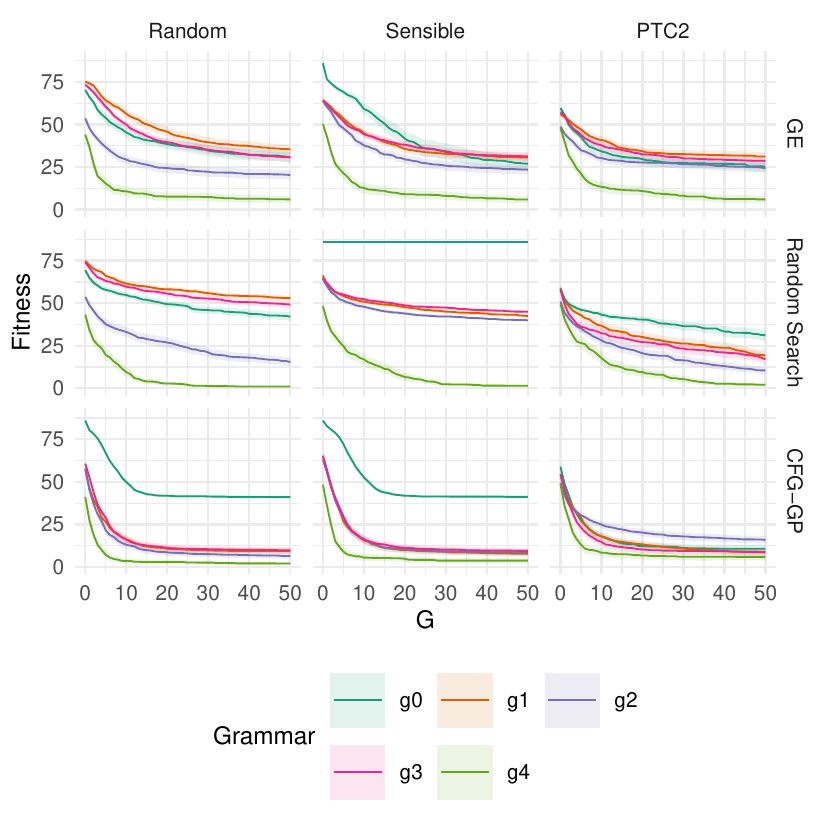}
  \caption{Fitness evolution over generations for the {\em Santa Fe Trail} problem.} \label{ant}
\end{figure}

All results in Figures~\ref{keijzer6}--\ref{ant-limits} are presented
with 95\% confidence intervals around the relevant statistic as shaded
regions. Out-of-sample test cases are also reported for the symbolic
regression problems, however our emphasis in this paper is to examine
the training (i.e., {\em search}) behaviour of the methods: while
generalisation performance of the evolved models is the primary
statistic of interest in terms of application, in this paper we make
no attempt to control overfitting through regularisation or other
means. Therefore, high test errors could be equally attributable to
either overfitting (i.e., very strong search during training) or
underfitting (i.e., weak search during training).

\begin{figure*}[t]
  \centering
  \begin{subfigure}{0.8\linewidth}
\begin{verbatim}
<code>         ::= <line> | <code> <line>
<line>         ::= <expr>
<expr>         ::= <if-statement> | <op>
<if-statement> ::= if (food_ahead()) <if-true> <if-false>
<op>           ::= left(); | right(); | move();
<if-true>      ::= { <expr> }
<if-false>     ::= else { <expr> }
\end{verbatim}
    %%\vspace{-1.5em}
    \caption{g0: used by O'Neill and Ryan~\cite{o1999evolving}}
  \end{subfigure}\\[1em]
  \begin{subfigure}{0.8\linewidth}
\begin{verbatim}
<expr>         ::= <line> | <expr> <line>
<line>         ::= if (food_ahead()) { <expr> } else { <expr> } | <op>
<op>           ::= left(); | right(); | move();
\end{verbatim}
    %%\vspace{-1.5em}
    \caption{g1: used by Robilliard et al.~\cite{robilliard2005santa}}
  \end{subfigure}\\[1em]
  \begin{subfigure}{0.8\linewidth}
\begin{verbatim}
<code>         ::= <line> | <code> <line>
<line>         ::= <if> | <op>
<if>           ::= if (food_ahead()) { <line> } else { <line> }
<op>           ::= left(); | right(); | move();
\end{verbatim}
    %%\vspace{-1.5em}
    \caption{g2: used by Whigham et al.~\cite{whigham2015examining} and previously by
      Nicolau et al. to emphasise incorrect grammar design~\cite{nicolau2012termination}}
  \end{subfigure}\\[1em]
  \begin{subfigure}{0.8\linewidth}
\begin{verbatim}
<code>         ::= <expr> | <code> <line>
<expr>         ::= <line> | <line> <expr>
<line>         ::= if (food_ahead()) { <expr> } else { <expr> } | <op>
<op>           ::= left(); | right(); | move();
\end{verbatim}
    %%\vspace{-1.5em}
    \caption{g3: used by Harper~\cite{harper2010ge}}
  \end{subfigure}\\[1em]
  \begin{subfigure}{0.8\linewidth}
\begin{verbatim}
<prog>         ::= <c> <c> <c> <c> <c> <c> <c> <c>
<c>            ::= <line> | '' ## the '' emulates a no-op
<line>         ::= if (food_ahead()) <op> else <op>
<op>           ::= left(); | right(); | move();
\end{verbatim}
    %%\vspace{-1.5em}
    \caption{g4: designed through analysis of solutions generated using grammar g0}
  \end{subfigure}
  \caption{The grammars used for the {\em Santa Fe Trail} problem.}
  \label{fig:ant-grammars}
\end{figure*}

The results of experiments on the {\em Keijzer-6} problem are shown in
Figure~\ref{keijzer6}. Confirming the results of previous work, GE is
particularly sensitive to both initialisation and grammar design on
this problem. Interestingly, the behaviour of random search is
characteristically similar to that of GE, while the behaviour of
CFG-GP on this problem appears to be largely unaffected by the choice
of initialisation. CFG-GP also appears to be somewhat more robust to
changes to the grammar than both GE and random search. The test
performances of the best-trained individuals for this problem are
presented in Figure~\ref{keijzer6-test} and are generally consistent
with the trend observed during the training phase.

The results of experiments on the {\em Vladislavleva-4} problem are
shown in Figures~\ref{vladislavleva4}
and~\ref{vladislavleva4-test}. Compared to the {\em Keijzer-6}
problem, there is far more consistency in training across all methods
on this problem. However, there appears to be a slight increase in
performance by using CFG-GP in later generations. Across all three
methods, there seems to be little impact on performance (in terms of
final solution quality) by changing the initialisation method. On
inspection of the solutions produced by GE and Random Search, they
tend to be very short (no more than three terminals) and are
discovered early in the run. In contrast, solutions CFG-GP tend to be
larger and more expressive: however, the grammar in this problem tends
to encourage the unconstrained use of power functions, which means
that very high test errors are quickly obtained frequently,
particularly so for CFG-GP.

Next, we examine the four symbolic regression problems, the results of
which are presented in Figures~\ref{regression}
and~\ref{regression-test}. The intention of these experiments was to
examine the theory that reducing non-terminals in the grammar is
beneficial to performance. Interestingly, the only problem where this
theory is possibly observed is in the {\em Forest Fires} problem,
although the differences observed between grammars are not
statistically significant. Interestingly, CFG-GP seems to be the
method that benefits from non-terminal reduction the most, whereas on
at least the {\em Boston Housing} and {\em Tower} problems, both GE
and random search seemed to benefit from the added structure provided
by the additional non-terminals in the grammar. In both of these
cases, GE and random search appear to find good solutions faster using
the structured g0 grammar over the compromise g6 grammar. In previous
work, non-terminal reduction and bias reduction coincided with a shift
from infix to prefix notation, whereas in this work both g0 and g6
grammars use a prefix notation. It is plausible that the change in
notation was the motivator for improved performance rather than the
other grammar modifications. Aside from observation, it is also
noteworthy that all three methods appear to be insensitive to
initialisation on these problems, and that again, the performance of
GE is very similar to that of random search on all but the {\em Tower}
problem. The test performance on these problems again shows that
CFG-GP behaviour is fairly consistent in the presence of changes to
initialisation and grammar: generally, CFG-GP offers the best testing
performance on three of the four problems, with test performance on
the {\em Forest Fires} being similar to that of {\em Vladislavleva-4}
(CFG-GP errors are due to overfitting, while GE and Random Search
exhibit underfitting).

The results of experiments on the {\em Shape} problem are shown in
Figure~\ref{shape}. The results here present an interesting behaviour
for CFG-GP, which struggles to find solutions when using the g5 and g6
grammars. As discussed in the next section, this appears to be an
interaction between the depth limiting used in CFG-GP and depths of
the derivation trees that these grammars need to describe highly fit
solutions. In contrast to CFG-GP, initialisation seems to play an
important role in the behaviour of random search and GE on this
problem.

Finally, Figure~\ref{ant} shows the evolution of fitness on the {\em
  Santa Fe Trail} problem. Again, an interesting observation is made
of the performance of CFG-GP on this problem: when using the g0
grammar, search appears to stagnate quickly resulting in poor
fitness. A similar observation is made for random search on this
problem. As discussed in the next section, this again appears to be a
consequence of parameter choice in CFG-GP, this time in the maximum
depth of the trees generated in subtree mutation. For the remaining
grammars, there is a clear advantage to using CFG-GP over the other
two methods. Most interesting, however, is the g4 grammar that was
developed specifically for this paper: this grammar was not designed
or adapted to better align with any specific representation, but
rather to bias search to encourage the discovery of better
solutions. As can be seen, using this grammar leads to effective
search regardless of chosen method.

\section{Discussion}\label{sec:discussion}
The results in the previous section suggest that two interesting
outcomes:
\begin{enumerate}
\item in agreement with previous work, there seems to be a closer
  relationship between the behaviour of random search and GE than
  there is between random search and CFG-GP, as confirmed by the
  results on the {\em Keijzer-6}, {\em Vladislavleva-4}, {\em Boston
    Housing}, {\em Dow Chemical} and {\em Forest Fire} problems; and
\item in general, CFG-GP seems reasonably robust to the choice of
  initialisation method, and is able to perform a good search
  following varying degrees of initial search quality.
\end{enumerate}
While CFG-GP appears to offer the overall most stable and effective
search performance, its behaviour on the {\em Shape} problem (using
grammars g5 and g6) and the {\em Santa Fe Trail} (using grammar g0) is
poor. However, an analysis of the grammars, and the requirements of
these grammars in terms of the derivation trees required to produce
good solutions yields useful information that can help the performance
of CFG-GP on these problems. In the case of the {\em Shape} problem,
examples of optimal solutions are provided in previous
work~\cite{o2009shape}. The derivation trees for these optimal solutions
have a depth of around 20, which is outside the depth limits imposed
on CFG-GP in this work. Similarly, for the {\em Santa Fe Trail}, the
production that ultimately produces {\tt if (food\_ahead())}\ldots{}
expressions requires a minimum derivation tree depth of 5 to be fully
terminated. Given that CFG-GP operated with a maximum mutation depth
of 4, this meant that no such expressions could appear in solutions
through mutation, which greatly inhibited search quality on this
problem using this grammar.

\begin{figure}[t]
  \centering
  \includegraphics[width=\linewidth]{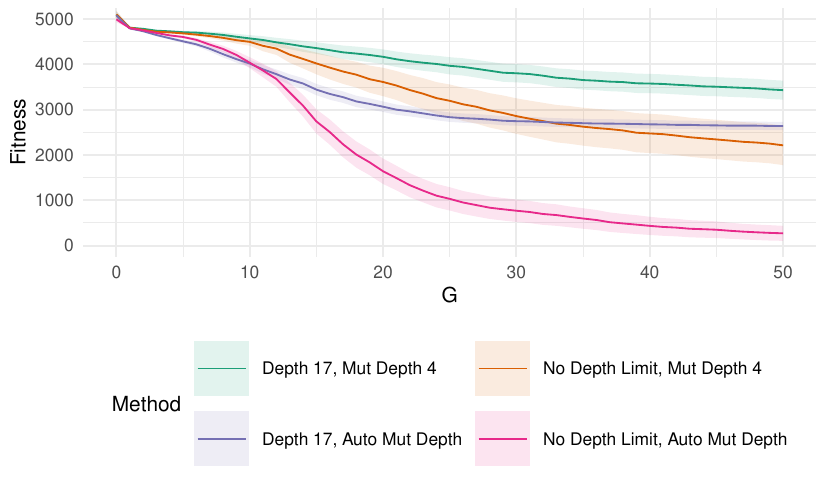}
  \caption{Effect of depth limiting and mutation depth on the {\em Shape} problem.} \label{shape-limits}
\end{figure}

\begin{figure}
  \centering
  \includegraphics[width=\linewidth]{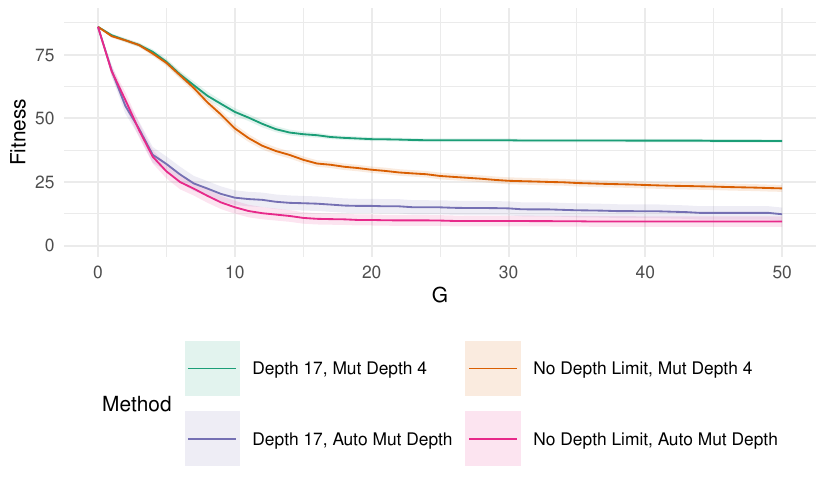}
  \caption{Effect of depth limiting and mutation depth on the {\em Santa Fe Trail} problem.}\label{ant-limits}
\end{figure}

Given the insights developed from examining the grammars for the {\em
  Shape} and {\em Santa Fe Trail} problems, a simple modification to
CFG-GP was proposed. First, depth limiting was removed, and then a
small modification was made to subtree mutation: rather than setting a
hard limit, maximum mutation depth was computed as the larger of the
either the depth of the largest production for the non-terminal being
mutated, or the depth of the subtree that mutation is replacing. The
results of this modified configuration of CFG-GP are shown in
Figures~\ref{shape-limits} and \ref{ant-limits}. All results are
presented using the g6 grammar for {\em Shape} and the g0 grammar for
{\em Santa Fe Trail}. As can be seen, the simple modifications to
CFG-GP have markedly improved performance.

The solution proposed here for CFG-GP is simple, but appears to be
effective. It must be acknowledged that the insight that led to
removing or increasing depth limits on the {\em Shape} problem
required knowledge of the optimal solutions for this problem. However,
the proposed solution shed light new light on the understanding of the
tuning parameters of CFG-GP. Traditionally, CFG-GP has been configured
using similar parameter settings to that of standard genetic
programming; it is likely that, due to the shift from parse trees to
derivation trees, the parameter values typically used for standard genetic
programming are not suitable for CFG-GP. Future work should be
dedicated to developing a more thorough understanding of the tuning
parameters of CFG-GP independent from any understanding of standard
genetic programming.

\section{Conclusion}\label{sec:conc}
Grammars provide a simple and powerful method of incorporating
structure and domain knowledge into evolutionary search. Recent work
suggests that grammatical evolution, currently the most popular form
of grammar-guided evolutionary computation, requires extensive tuning
to problems through careful grammar design and is heavily reliant upon
accurate initialisation of the population for good
performance. However, the behaviour of these modifications to GE has
not been developed in reference to other methods of grammar-guided
search, such as context-free grammar genetic programming. Hence the
relative merits of these modifications has been an unknown
quantity. This paper attempts to shed light of the relative merit of
these modifications to GE by repeating similar experiments using
random search and CFG-GP as a reference point. When tested on several
benchmark problems, it is shown that the modified variants of GE do
demonstrate improved performance, but this performance is still often
inferior to that of CFG-GP. Additionally, the results presented here
suggest that CFG-GP is less sensitive to the effects of initialisation
and the design of the grammar. Finally, the results presented here
also suggest that CFG-GP may be able to adapt to more problems through
simple tuning of its parameters, rather than resorting to more
elaborate modifications to the grammar.

The results presented here suggest several areas of future
work. Clearly, the parameter space of CFG-GP is not as
fully-understood as it could be, so future work should develop a
better understanding of how the parameters of CFG-GP impact its
behaviour. Likewise, there would be significant benefit from future
work developing new methods for grammar design that emphasise {\em
  more effective expression of the problem being searched} rather than
focusing on {\em distorting the grammar to fit the chosen
  representation}.

\bibliographystyle{ACM-Reference-Format}
\bibliography{grammar-design}

\end{document}

%% file: abstract.tex
%% Grammars provide a convenient and powerful mechanism to define the
%% space of possible solutions for a range of problems. However, when
%% used in grammatical evolution (GE), great care must be taken in the
%% design of a grammar to ensure that the polymorphic nature of the
%% genotype-to-phenotype mapping does not impede search. Additionally,
%% recent work has highlighted the importance of the initialisation
%% method on GE's performance. While recent work has shed light on the
%% matters of initialisation and grammar design with respect to GE, their
%% impact on other methods, such as random search and context-free
%% grammar genetic programming (CFG-GP), is largely unknown. This paper
%% examines GE, random search and CFG-GP under a range of benchmark
%% problems using several different initialisation routines and grammar
%% designs. The results suggest that CFG-GP is less sensitive to
%% initialisation and grammar design than both GE and random search. In
%% addition, any poor performance observed by CFG-GP appears to be
%% managed through simple adjustment of tuning parameters, rather than
%% resorting to extensive modification of the grammar to fit the search
%% representation. We conclude that CFG-GP is a stronger base than GE
%% from which to conduct grammar-guided evolutionary search, and that
%% future work should focus on understanding the parameter space of
%% CFG-GP for better application.
Grammars provide a convenient and powerful mechanism to define the
space of possible solutions for a range of problems. However, when
used in grammatical evolution (GE), great care must be taken in the
design of a grammar to ensure that the polymorphic nature of the
genotype-to-phenotype mapping does not impede search. Additionally,
recent work has highlighted the importance of the initialisation
method on GE's performance. While recent work has shed light on the
matters of initialisation and grammar design with respect to GE, their
impact on other methods, such as random search and context-free
grammar genetic programming (CFG-GP), is largely unknown. This paper
examines GE, random search and CFG-GP under a range of benchmark
problems using several different initialisation routines and grammar
designs. The results suggest that CFG-GP is less sensitive to
initialisation and grammar design than both GE and random search: we
also demonstrate that observed cases of poor performance by CFG-GP are
managed through simple adjustment of tuning parameters. We conclude
that CFG-GP is a strong base from which to conduct grammar-guided
evolutionary search, and that future work should focus on
understanding the parameter space of CFG-GP for better application.